\documentclass{article}

\usepackage[preprint]{neurips_2021}

\usepackage[utf8]{inputenc} 
\usepackage[T1]{fontenc}    
\usepackage{hyperref}       
\usepackage{url}            
\usepackage{booktabs}       
\usepackage{amsfonts}       
\usepackage{nicefrac}       
\usepackage{microtype}      
\usepackage{xcolor}         
\usepackage{graphicx}
\hypersetup{
    colorlinks = true,
    linkbordercolor = {white},
    citecolor={brown}
}
\usepackage{amsmath,amsfonts,amssymb,amsthm, bm}

\title{A Note on Connecting Barlow Twins with
\\Negative-Sample-Free Contrastive Learning}

\author{%
  Yao-Hung Hubert Tsai, Shaojie Bai, Louis-Philippe Morency, Ruslan Salakhutdinov \\
  Machine Learning Department, Carnegie Mellon University
}

\begin{document}

\maketitle

\vspace{-4mm}

Barlow Twins~\citep{zbontar2021barlow} is a recently proposed self-supervised learning (SSL) method that encourages similar representations between distorted variations (augmented views) of a sample, while minimizing the redundancy \emph{within} the representation vector. Specifically, compared to the prior state-of-the-art SSL methods, Barlow Twins demonstrates two main properties. On one hand, its algorithm requires no explicit construction of negative sample pairs, and is not sensitive to large training batch sizes, both of which are the characteristics commonly seen in the recent \emph{non-contrastive} SSL methods (e.g., BYOL~\citep{grill2020bootstrap} and SimSiam~\citep{chen2020exploring}).  On the other hand, it avoids the reliance on symmetry-breaking network designs for distorted samples, which had been found to be crucial for these non-contrastive approaches (in order to avoid learning collapsed representations). We note that 
designing symmetry-breaking networks is not needed for the recent \emph{contrastive} SSL methods (e.g., SimCLR~\citep{chen2020simple})\footnote{Another SSL approach is the MoCo~\citep{he2020momentum} method. It belongs to the class of contrastive approaches since it requires constructing negative samples. Compared to standard contrastive approaches, such as SimCLR~\citep{chen2020simple}, it additionally introduces symmetry-breaking network design, which enables MoCo to be robust to various batch-sizes. In contrast, the Barlow Twins method avoids the need for the symmetry-breaking network design, while still being robust to various batch-sizes.}. A natural question arises: \emph{what makes Barlow Twins an outlier among the existing SSL algorithms}?

\citet{zbontar2021barlow} motivated the algorithmic design of Barlow Twins via the Information Bottleneck (IB) theory~\citep{tishby2000information}, assuming the representations are Gaussian and that we need to minimize the determinant of the feature cross-correlation. In this note we provide an alternative interpretation of the Barlow Twins' objective by 
viewing it as a \emph{negative-sample-free} contrastive learning objective. Specifically, we relate the method to Hilbert-Schmidt Independence Criterion (HSIC)~\citep{gretton2005measuring} maximization between augmented views, where HSIC is a contrastive approach that avoids the need of constructing negative pairs. We show that this new interpretation entails no assumption on the Gaussianity, and is thus more consistent with how Barlow Twins is used in practice. While the HSIC interpretation calls for a slight change to the original Barlow Twins' objective, we empirically show that the change leads to no loss in performance\footnote{ See~\url{https://github.com/yaohungt/Barlow-Twins-HSIC} for the implementation details.}. 

Overall, we believe that such \emph{negative-sample-free} contrastive approach (exemplified by Barlow Twins)  can serve as a bridge between the two major families of SSL methods: non-contrastive (e.g., BYOL~\citep{grill2020bootstrap},  SimSiam~\citep{chen2020exploring}) and  contrastive approaches (e.g., SimCLR~\citep{chen2020simple}).

\section{Barlow Twins' Method}

We first obtain the distorted variations of a given sample by applying different augmentations (i.e., Gaussian blurring, rotation, translation, etc.). Then, Barlow Twins' method~\citep{zbontar2021barlow} encourages the empirical cross-correlation matrix (between learned representations of the distorted variations of a given sample) to be an identity matrix.  The rationale is that 1) when on-diagonal terms of the cross-correlation matrix take value $+1$, it encourages the learned representations of distorted versions of a sample to be similar; and 2) when off-diagonal terms of the cross-correlation matrix take value $0$, it encourages the diversity of learned representations, since different dimensions of these representations will be uncorrelated to each other.

For notations simplicity, throughout the report, we assume $X \in \mathbb{R}^{n\times d}$ and $Y \in \mathbb{R}^{n\times d}$ to be standardized features (mean zero and standard deviation one) from the distorted versions of a given sample. To be more precise, 
$$X={\rm Standardization\Big(Feature\_Encoder\big(Augmented\_Data\_1\big)\Big)}$$ and $$Y={\rm Standardization\Big(Feature\_Encoder\big(Augmented\_Data\_2\big)\Big)}.$$ $n$ is the number of samples and $C$ = $ \frac{X^\top Y}{n} \in \mathbb{R} ^{d\times d}$ is the empirical cross-correlation matrix. Barlow Twin's objective considers the following loss:
\begin{equation}
    {\rm Barlow\,\,Twins'\,\,Loss\,\,} = \,\sum_{i} (1 - C_{ii})^2 + \lambda \sum_i \sum_{j \neq i} C_{ij}^2,
\label{eq:barlow}
\end{equation}
where $\lambda$ is a trade-off parameter controlling the on- (i.e., $C_{ii}$) and off-diagonal terms (i.e., $C_{ij}$ for $i\neq j$). In \citet{zbontar2021barlow}, the authors find that $\lambda = 0.005$ works the best using a grid search.

\section{Hilbert-Schmidt Independence Criterion for Self-supervised Learning}

We denote $P_{XY}$ and $P_XP_Y$ as the joint and the product of marginal distributions of the features $X$ and $Y$. Contrastive learning objectives~\citep{chen2020simple,hjelm2018learning,ozair2019wasserstein,tsai2021self} aim to learn similar representations for positively paired data (i.e., $x$ would be similar to $y$ when $(x,y)\sim P_{XY}$) and learning dissimilar representations for negatively paired data (i.e., $x$ would be dissimilar to $y$ when $(x,y)\sim P_{X}P_{Y}$). For example, $ P_{XY}$ denotes the distribution of learned representations from distorted variations of a given sample and $P_{X}P_{Y}$ denotes the distribution of learned representations from different samples. \citet{tsai2021self} showed that the contrastive objectives can be interpreted as maximizing the probability divergence between $P_{XY}$ and $P_{X}P_{Y}$, where the probability divergence can be either Kullback–Leibler  divergence~\citep{chen2020simple}, Jensen–Shannon divergence~\citep{hjelm2018learning}, Wasserstein divergence ~\citep{ozair2019wasserstein} or $\chi^2$ divergence~\citep{tsai2021self}. Hilbert-Schmidt Independence Criterion (HSIC)~\citep{gretton2005measuring} also represents a probability divergence between $P_{XY}$ and $P_{X}P_{Y}$, where it considers the maximum mean discrepancy (MMD)~\citep{gretton2012kernel} as the probability divergence measurement. In particular, 
\begin{equation}
{\rm HSIC} (X,Y) := {\rm MMD}\, (\, P_{XY} \,\|\, P_{X}P_Y\,) =  \| C_{XY} \|_{\rm HS}^2,
\label{eq:HSIC_mmd}
\end{equation}
where $C_{XY}$ is the cross-covariance operators between the Reproducing Kernel Hilbert Spaces (RKHSs) of $X$ and $Y$ and  $\| \cdot \|_{\rm HS}^2$ is the Hilbert-Schmidt norm. \citet{gretton2005measuring} proposed an empirical estimation of HSIC:
\begin{equation}
\widehat{{\rm HSIC} (X,Y)} := \frac{1}{n^2}\,{\rm tr}\,\Big({\bf K_X H K_Y H} \Big),
\label{eq:HSIC_emp}
\end{equation}
where ${\bf K_X}$ and ${\bf K_Y}$ are kernel Gram matrices of $X$ and $Y$ respectively and ${\bf H}  = {\bf I} - \frac{1}{n} {\bf 1}{\bf 1}^\top \in \mathbb{R}^{n\times n}$ is the centering matrix. We note that the above equation only requires the positively paired samples to construct the kernel ${\bf K_X}$ and ${\bf K_Y}$. Hence, adopting HSIC as a contrastive learning objective {\em avoids the need of} explicitly 
minimizing the similarities between the negatively paired data, which is crucial in prior contrastive learning methods~\citep{chen2020simple,hjelm2018learning,ozair2019wasserstein,tsai2021self}. In short, HSIC objective can be interpreted as a {\em negative-sample-free} contrastive learning objective. In what follows, we will show that by considering the linear kernel as the characteristic kernel in the RKHSs, we can relate HSIC objective to the Barlow Twins' objective.

\subsection{Connecting HSIC to Barlow Twins}
To connect HSIC to Barlow Twins, we let the characteristic kernel in the RKHSs to be the linear kernel, in particular ${\bf K_X} = X X^\top$ and ${\bf K_Y} = Y Y^\top$. Since $X$ and $Y$ are standardized (mean $0$), it is not hard to show that $X X^\top {\bf H} = X X^\top$ and $Y Y^\top {\bf H} = Y Y^\top$. Plugging in this result into equation~\eqref{eq:HSIC_emp}, we obtain:
\begin{equation}
    \begin{split}
    \widehat{{\rm HSIC} (X,Y)} := & \frac{1}{n^2}\,{\rm tr}\,\Big({\bf K_X H K_Y H} \Big) = \frac{1}{n^2} \,{\rm tr}\,\Big(X X^\top Y Y^\top \Big) \\ 
    = & \frac{1}{n^2} \,{\rm tr}\,\Big(X^\top Y \big( X^\top Y \big)^\top \Big)  = \frac{1}{n^2} \| X^\top Y \|_F^2 = \| C \|_F^2,
    \end{split}
    \label{eq:hsic_cross_correlation}
\end{equation}
where $C$ is the empirical cross-correlation matrix by definition and $\|\cdot\|_F^2$ is the Frobenius norm. We see that, when considering the linear kernel (i.e., $k(x,y) = \langle\, x, y\rangle$), the Hilbert-Schmidt norm of the cross-covariance operator in equation~\eqref{eq:HSIC_mmd} equals to the Frobenius norm of the cross-covariance matrix. And we note that, since $X$ and $Y$ are standardized, the cross-covariance matrix becomes the cross-correlation matrix, where $C_{ij} \in [-1, 1]\,\, \forall\, i, j \in \{1, \cdots, d\}$.

Contrastive learning approaches aim to maximize the distribution divergence between $P_{XY}$ and $P_XP_Y$, which means that we can maximize equation~\eqref{eq:hsic_cross_correlation} for self-supervised representation learning. Nonetheless, a trivial solution $C_{ij} = 1\,\, \forall\, i, j \in \{1, \cdots, d\}$ exists when maximizing equation~\eqref{eq:hsic_cross_correlation}, which is not ideal since the perfect correlation ($+1$) between different dimensions of the representations implies a low power of the representations. By noticing that $C_{ij} = +1$ or $C_{ij} = -1$ both maximizes $C_{ij}^2$, we can prevent this trivial solution by encouraging 1) the on-diagonal terms of $C$ to be $+1$ and 2) the off-diagonal terms of $C$ to be $-1$. Then, the resulting loss is:
\begin{equation}
    {\rm HSIC_{\rm ssl}\,\,} = \,\sum_{i} (1 - C_{ii})^2 + \lambda \sum_i \sum_{j \neq i} (1 + C_{ij})^2,
\label{eq:hsic_ssl_loss}
\end{equation}
where $\lambda$ is a trade-off hyper-parameter. Equation~\eqref{eq:hsic_ssl_loss} resembles the original Barlow Twins' objective (equation~\eqref{eq:barlow}) with the difference on the off-diagonal terms: Barlow Twins encourages the off-diagonal terms to be {\em zero} and the ${\rm HSIC_{\rm SSL}}$ encourages the off-diagonal terms to be {\em negative one}. 

\subsection{Discussion} 

{\it Connecting ${\rm HSIC_{\rm SSL}}$ to downstream tasks}: So far we see that minimizing ${\rm HSIC_{\rm SSL}}$ (equation~\eqref{eq:hsic_ssl_loss}) encourages the dependency between learned representations from the distorted samples (i.e., $X$ and $Y$). Maximizing this dependency can be viewed as maximizing the mutual information between learned representations, where~\citet{tsai2021multiview} showed that this process acts to {\em extract downstream task-relevant information}.  A complementary objective to extracting downstream-task-relevant information is the objective that {\em discards downstream-task-irrelevant information}, such as the squared loss between $X$ and $Y$ (i.e., $\| X-Y \|_F^2$, see \citet{tsai2021multiview}). We will show that minimizing the squared loss between $X$ and $Y$ equals to encouraging the on-diagonals of $C$ to be $+1$:
\begin{equation}
    \begin{split}
    \|X - Y \|_F^2 & = {\rm tr} \Big( \big(X - Y\big)^\top \big(X - Y\big) \Big) = {\rm tr} \Big( X^\top X + Y^\top Y - X^\top Y - Y^\top X \Big) \\
    & = 2d - 2 {\rm tr} \Big( X^\top Y\Big) = 2d - 2n\, {\rm tr} \Big( C\Big),
    \end{split}
\end{equation}
where $X^\top X = Y^\top Y = d$ because $X$ and $Y$ are standardized representations. Since the cross-correlation ranges between $-1$ and $+1$, maximizing ${\rm tr} \,\big( C \big)$ encourages the on-diagonal entries of $C$ to be $+1$. To conclude, ${\rm HSIC_{\rm SSL}}$ (equation~\eqref{eq:hsic_ssl_loss}) plays the role to both {\em extract downstream-task-relevant} (i.e., maximizing the mutual information between $X$ and $Y$) and {\em discard downstream task-irrelevant information} (i.e., minimizing the squared loss between $X$ and $Y$).

{\it The choice of $\lambda$:} As suggested by~\citet{zbontar2021barlow}, one can find $\lambda$ by a grid search. Another choice is to set $\lambda=\frac{1}{d}$ since we want to balance on-diagonal and off-diagonal terms in equation~\eqref{eq:hsic_ssl_loss}. Note that equation~\eqref{eq:hsic_ssl_loss} contains $d$ on-diagonal terms and $d\cdot (d-1)$ off-diagonal terms, and the ratio is $\frac{1}{d-1} \approx \frac{1}{d}$. We empirically find this choice of $\lambda$ to work well when altering the dimension $d$, either for the Barlow Twins' method (equation~\eqref{eq:barlow}) or ${\rm HSIC_{\rm SSL}}$ (equation~\eqref{eq:hsic_ssl_loss}).

\section{Experiments}

The experiments considered in this technical report serves to fairly compare the Barlow Twins' loss (equation~\eqref{eq:barlow}) and ${\rm HSIC_{\rm SSL}}$ (equation~\eqref{eq:hsic_ssl_loss}), where we will later show that there is negligible performance difference between these two objective functions. Throughout the experiments, we set $\lambda = \frac{1}{d}$ in both the Barlow Twins' loss and ${\rm HSIC_{\rm SSL}}$ with $d$ being the dimension of the features. We choose CIFAR-10~\citep{krizhevsky2009learning} and Tiny-Imagenet~\citep{le2015tiny} as our datasets. 

\paragraph{Training and Evaluation.}
We choose ResNet-50~\citep{he2016deep} as our backbone model (serving as the {\em encoder} described in~\citep{chen2020simple}) and replace the last classification layer with a non-linear projection layer (serving as the {\em projection head} described in~\citep{chen2020simple}). The encoder takes the input of an image and outputs a $2048$-dimensional feature. The projection head takes the input of the $2048$-dimensional feature and outputs a $d$-dimensional feature. Both the Barlow Twins's loss and ${\rm HSIC_{\rm SSL}}$ consider the $d$-dimensional features as $X$ and $Y$. During training, we update the parameters in the encoder and the projection head to minimize the chosen loss, $1,000$ epochs on CIFAR10 and $650$ epochs on Tiny ImageNet. The inputs are distorted variations of input samples. During evaluation, we remove the projection head and report the metric on the test images using the $2048$-dimensional feature with the linear classification~\citep{chen2020simple} accuracy. The linear classification approach categorizes test images by additionally training another linear classifier on top of the $2048$-dimensional features, $200$ epochs for both datasets. More details can be found in~\url{https://github.com/yaohungt/Barlow-Twins-HSIC}.

\begin{figure}[t!]
\begin{center}
\includegraphics[width=0.8\textwidth]{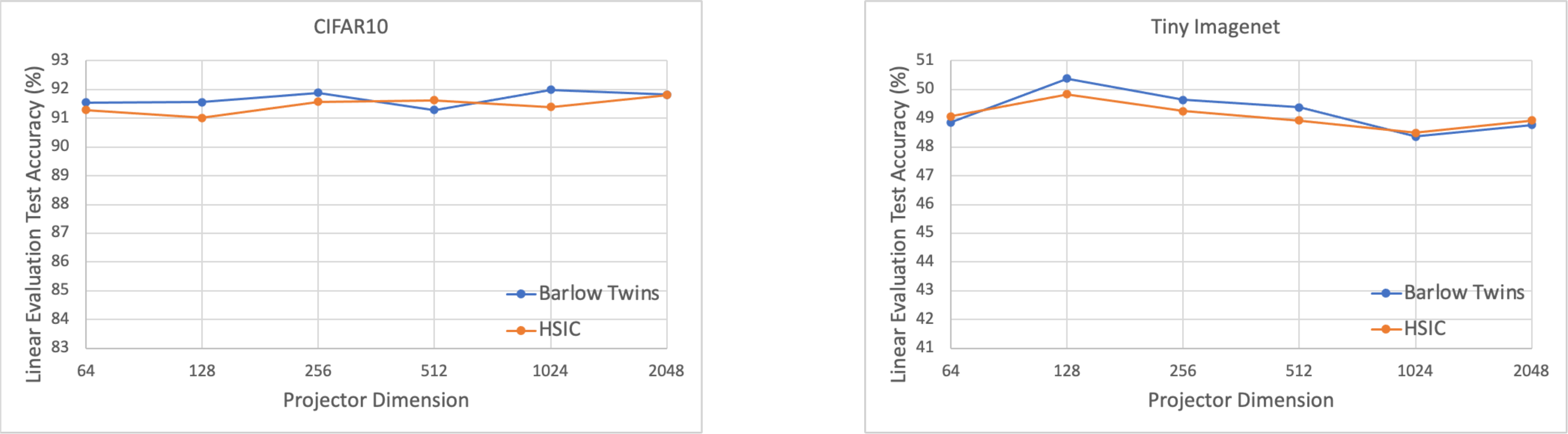}
\end{center}
\vspace{-4mm}
\caption{\small Comparisons of the learned self-supervised representations on CIFAR10 and Tiny ImageNet by changing the projector dimension $d$. The batch size is fixed to $128$ and $\lambda$ is set to be $\frac{1}{d}$.}
\label{fig:proj_dim}
\end{figure}

\paragraph{Results and Discussions.}
First, in Figure~\ref{fig:proj_dim}, we compare the learned self-supervised representations on CIFAR10 and Tiny ImageNet by changing the projector dimension $d$. The batch size is fixed to $128$ and $\lambda$ is set to be $\frac{1}{d}$. We find that the performance difference between Barlow Twins and ${\rm HSIC_{\rm SSL}}$ is negligible. We also find that the performance does not differ much when altering $d$. This observation has a conflict with the original Barlow Twins' paper~\citep{zbontar2021barlow}, where it states that the Barlow Twins' method performs better when the projector dimension $d$ is large \Big(see its Figure 4\Big). We argue that this disaccordance is because of 1\Big) the self-supervised learned representations behave differently on large datasets \Big(ImageNet considered by~\citet{zbontar2021barlow}\Big) and small datasets \Big(CIFAR10 and Tiny ImageNet considered by us\Big); and 2\Big) the choice of $\lambda$ \Big(we set $\lambda = \frac{1}{d}$, and it is unclear that has $\lambda$ be chosen by extensively grid search in Figure 4 in the paper~\citep{zbontar2021barlow}\Big).

\begin{figure}[t!]
\begin{center}
\includegraphics[width=0.8\textwidth]{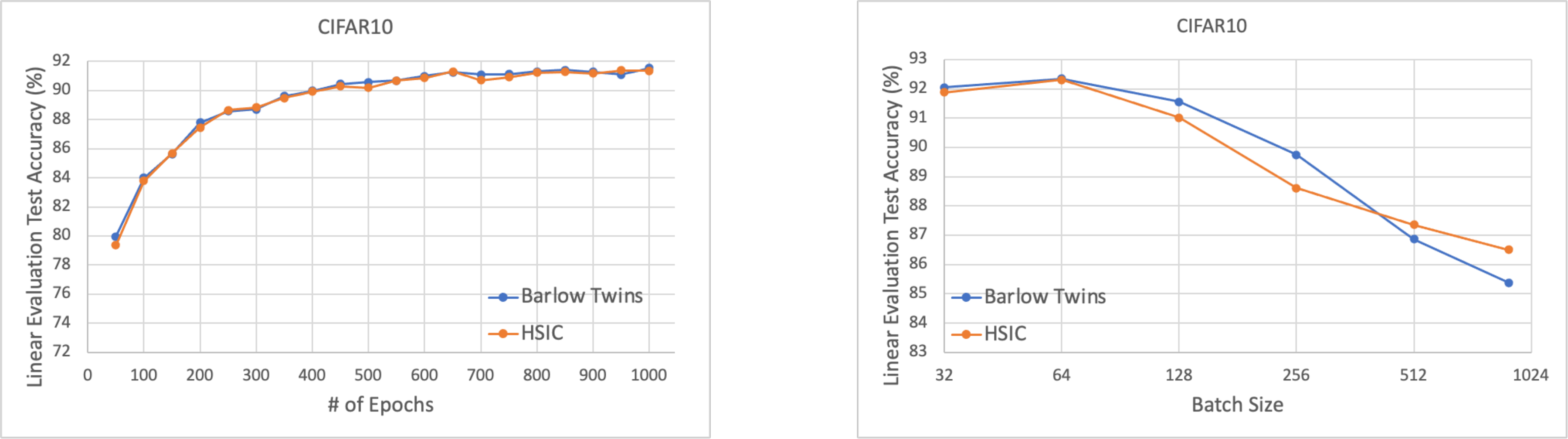}
\end{center}
\vspace{-4mm}
\caption{\small Comparisons of the learned self-supervised representations on CIFAR10 by changing the number of training epochs and effect of batch size. The projector dimension is fixed to $128$ and $\lambda$ is set to be $\frac{1}{128}$.}
\label{fig:epoch_bsz}
\end{figure}

Second, in Figure~\ref{fig:epoch_bsz}, we compare the learned self-supervised representations on CIFAR10 by changing the number of training epochs and effect of batch size. The projector dimension is fixed to $128$ and $\lambda$ is set to be $\frac{1}{128}$. The same trend is observed: we do not find obvious performance difference between Barlow Twins and ${\rm HSIC_{\rm SSL}}$. Another interesting observation is that Barlow Twins and ${\rm HSIC_{\rm SSL}}$ reach to worsened performance when increasing the batch size. This observation has also be found in Figure 2 in the original Barlow Twins' paper~\citep{zbontar2021barlow}. We do not have a good explanation for this phenomenon and will continue on investigating it.

\section{Conclusion} 
In this report, we relate the algorithmic design of Barlow Twins' method~\citep{zbontar2021barlow} to the Hilbert-Schmidt Independence Criterion (HSIC), thus establishing it as a contrastive learning approach that is \emph{free of negative samples}. Through this perspective, we argue that Barlow Twins (and thus the class of negative-sample-free contrastive learning methods) suggests a possibility to bridge the two major families of self-supervised learning philosophies: non-contrastive and contrastive approaches. In particular, Barlow twins exemplified how we could combine the best practices of both worlds: avoiding the need of large training batch size and negative sample pairing (like non-contrastive methods) and avoiding symmetry-breaking network designs (like contrastive methods).
To conclude, we believe this work sheds light on the advantage of exploring the directions of designing {\em negative-sample-free} contrastive learning objectives.

{
\small
\bibliography{ref}
\bibliographystyle{plainnat}
}

\end{document}